\documentclass[letterpaper, 10 pt, conference]{ieeeconf}
\IEEEoverridecommandlockouts

\usepackage[utf8]{inputenc}

\usepackage{mathbbol}

\usepackage{esvect}

\usepackage[percent]{overpic}

\usepackage{tabu}

\usepackage{multicol}

\usepackage{multirow}

\usepackage[overload]{empheq}

\usepackage{bm}

\usepackage{xcolor}
\definecolor{dg}{rgb}{0.1, 0.6, 0.2}       
\definecolor{b}{rgb}{0.0, 0.0, 1}          

\usepackage{epsfig}
\usepackage{float}
\usepackage{color}
\usepackage{url}

\usepackage{algorithmic}
\usepackage[ruled,linesnumbered,vlined]{algorithm2e}

\usepackage{amssymb, amsfonts}
\usepackage{amsmath}
\usepackage{mathrsfs}

\usepackage{amsthm}
\usepackage{mathtools}

\let\labelindent\relax
\usepackage{enumitem}       
\setenumerate[enumerate]{align=left}

\usepackage{graphicx}

\usepackage{stackengine}

\usepackage[flushleft]{threeparttable}

\usepackage{balance}

\newcommand{\ceil}[1]{\left\lceil   #1 \right\rceil}

\newcommand{\norm}[1]{\left\lVert#1\right\rVert}

\makeatletter
\newlength\tmp@\newlength\t@mp
\newcommand{\comp}[3]
  {\mathop{ \settowidth\tmp@{$\displaystyle\mathop{#1}^{#3}_{#2}$}
  \hbox to \tmp@{\hss \settowidth\t@mp{$\displaystyle #1$}\setlength\t@mp{.45\t@mp}
  $\displaystyle\mathop{#1}^{\hspace\t@mp #3}_{\hspace{-\t@mp}#2}$
  \hss} }}
\makeatother

\newcommand{\Int}[2]
{\int_{#1}^{#2}}

\DeclareMathOperator*{\argmin}{argmin}

\usepackage[bookmarks=true]{hyperref}

\hypersetup{
    colorlinks=true,
    linkcolor=blue,
    filecolor=blue,      
    urlcolor=blue,
    citecolor=blue,
}

\def\d{\delta}

\def\o{\omega}

\def\x{\textbf{x}}

\def\D{\Delta}

\def\R{\mathbb{R}}

\def\ll{\left}
\def\rr{\right}

\def\pos{p}

\def\accel{\mathbf{a}}

\newcommand{\fr}[1]{\texttt{#1}}

\newcommand{\vbf}[1]{{\bm{\mathbf{#1}}}}

\def\Jcb{\vbf{J}}

\def\bias{\mathbf{b}}

\def\rot{R}
\def\tf{T}

\def\SO{\mathrm{SO(3)}}

\def\SE{\mathrm{SE(3)}}

\def\Exp{\mathrm{Exp}}
\def\Log{\mathrm{Log}}

\def\ident{\mathbf{I}}

\def\Dt{ {\D t} }

\def\X{\mathcal{X}}

\def\Z{\mathcal{Z}}

\def\I{\mathcal{I}}

\def\M{\mathcal{M}}
\def\L{\mathcal{L}}

\def\angvel{\bm{\omega}}
\def\accel{\mathbf{a}}
\def\grav{\mathbf{g}}
\def\eulvec{\bm{\phi}}

\newtheorem{remark}{Remark}

\def\fB{\fr{B}}

\def\fW{\fr{W}}

\def\f{\vbf{f}}
\def\n{\vbf{n}}

\newcommand{\ul}[1]{\underline{#1}}
\newcommand{\tb}[1]{\textbf{#1}}

\newcommand{\tref}[1]{Tab. \ref{#1}}

\def\res{\vbf{r}}
\def\tf{\vbf{T}}
\def\rot{\vbf{R}}
\def\pos{\vbf{p}}

\def\Abf{\vbf{A}}
\def\dbf{\vbf{d}}
\def\Pbf{\vbf{P}}

\begin{document}

\title{\bf Eigen Is All You Need:
Efficient Lidar-Inertial Continuous-Time Odometry with Internal Association}

\author{
      Thien-Minh Nguyen, \IEEEmembership{Member,~IEEE},
      Xinhang Xu,
      Tongxing Jin,
      Yizhuo Yang,\\
      Jianping Li,
      Shenghai Yuan,
      Lihua Xie, \IEEEmembership{Fellow,~IEEE}
\thanks{All authors are with the School of Electrical and Electronic Engineering, Nanyang Technological University, Singapore 639798, 50 Nanyang Avenue. 
This research is supported by the National Research Foundation, Singapore under its Medium Sized Center for Advanced Robotics Technology Innovation. Corresponding author: Thien-Minh Nguyen (email: thienminh.npn@ieee.org).}
}

\maketitle

\thispagestyle{plain}
\pagestyle{plain}

\begin{abstract}

In this paper, we propose a continuous-time lidar-inertial odometry (CT-LIO) system named SLICT2, which promotes two main insights. One, contrary to conventional wisdom, CT-LIO algorithm can be optimized by linear solvers in only a few iterations, which is more efficient than commonly used nonlinear solvers.
Two, CT-LIO benefits more from the correct association than the number of iterations. Based on these ideas, we implement our method with a customized solver where the feature association process is performed immediately after each incremental step, and the solution can converge within a few iterations. Our implementation can achieve real-time performance with a high density of control points while yielding competitive performance in highly dynamical motion scenarios. We demonstrate the advantages of our method by comparing with other existing state-of-the-art CT-LIO methods. For the benefits of the community, the source code is released at \url{https://github.com/brytsknguyen/slict}.

\end{abstract}

\IEEEpeerreviewmaketitle

\section{Introduction}

\begin{figure}
    \centering
    \begin{overpic}[width=0.9\linewidth,
                   ]{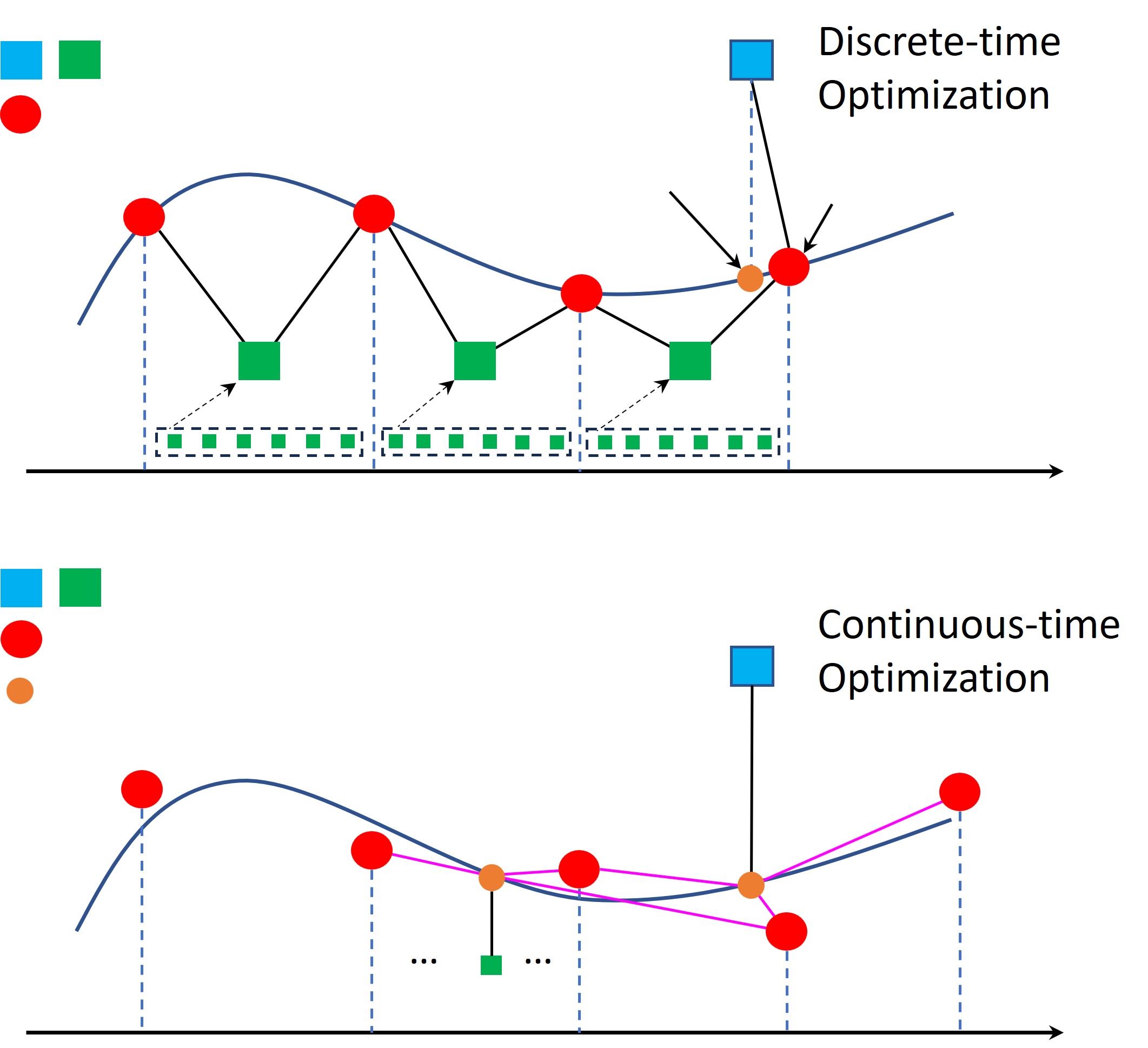}
                   \put(08, 45){ $t_w$}
                   \put(30, 45){ $t_{w+1}$}
                   \put(49, 45){ $\dots$}
                   \put(66, 45){ $t_{k}$}
                   \put(10, 85){\footnotesize lidar \& IMU mesurement}
                   \put(05, 80){\footnotesize discrete-time states}
                   \put(35, 75){\footnotesize true coupled state}
                   \put(69, 75){\footnotesize approx. coupled state}
                   \put(41, 55){\footnotesize preint.}

                   \put(10, -4){ $t_w$}
                   \put(30, -4){ $t_{w+1}$}
                   \put(50, -4){ $\dots$}
                   \put(66, -4){ $t_{k}$}
                   \put(80, -4){ $t_{k+1}$}
                   \put(10, 39){\footnotesize lidar \& IMU mesurement}
                   \put(05, 34){\footnotesize B-spline control points}
                   \put(05, 30){\footnotesize states at measurement times}
    \end{overpic}
\caption{Comparison between discrete-time (top) and continuous-time (bottom) optimization schemes. In the discrete case, measurements are coupled with distinct state estimates on a sliding window. Often, measurements are not acquired at the state's timestamp, thus some techniques are needed to enforce this strict state-measurement coupling paradigm. For example, the IMU measurements (small green squares) need to be preintegrated to provide a factor that couples two consecutive discrete states  (large green squares). On the other hand, for continuous-time optimization, measurements can be coupled directly with the trajectory estimate via the interpolated states. These interpolated states can be understood as time-weighted combination of the control points -- the actual decision variable in our optimization problem. In this illustration, we imply a spline order $N=3$, hence, each interpolated state is coupled with three control points (the pink segments).}
\label{fig: discrete vs continuous}
\vspace{-0.5cm}
\end{figure}
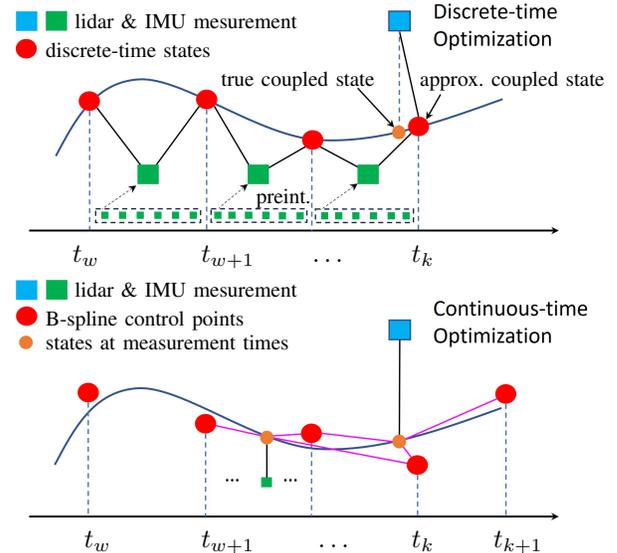

Continuous-time optimization has emerged as a new paradigm for lidar-based odometry and mapping (LOAM) systems in recent years. In general, the key aspect that differentiates continuous-time LOAM (CT-LOAM) from discrete-time systems is \textit{the use of raw lidar points in the optimization process, and the state estimates are formulated directly at the lidar point's sampling time}. On the contrary, in traditional discrete-time systems, only \textit{a number of state estimates at discrete time instances are created}, and some approximation technique is needed to convert the lidar point to a measurement that matches the state estimate's time (see the comparison in Fig. \ref{fig: discrete vs continuous}). The formulation of the instantaneous state estimate in continuous-time optimization can be based on piece-wise linear interpolation (PLI) \cite{dellenbach2022ct, nguyen2023slict} or B-spline polynomials \cite{quenzel2021real, lv2021clins, he2023continuous, zheng2024traj}.
Indeed, the B-spline formulation offers several advantages over the piece-wise linear model.
{Specifically, it can model the fast-changing trajectory better, fuse data of irregular sampling time, forgo IMU preintegration, and naturally supports online temporal offset estimation. However, it comes to our attention that the computational efficiency of CT-LOAM is still a bottleneck and has not been addressed adequately. This has motivated us to set out the investigation in this paper to obtain more insights on this issue}.

For some background, LOAM systems often employ some nonlinear solver (NLS) that optimizes a least-squares cost function by the Gauss-Newton method, of which Ceres-GTSAM-g2o \cite{Ceres-solver, GTSAM, grisetti2011g2o} is the dominant trio. Fundamentally, these solvers aim to solve a large least squares problem, in which individual square error terms are constructed and declared locally. The use of NLS offers several advantages. Most importantly, they allow one to tackle the problem in a divide-and-conquer manner, in which user only needs to focus on describing the residual and Jacobian of the individual factors with the locally-coupled states, whereas the arrangement of these residuals and Jacobians at the global scope is automatically managed by the solver. In addition, the NLS can provide extra features such as robust function to suppress outliers, options to exploit the sparseness of the problem, and in-depth diagnosis of the optimization process.

We also note that among the solvers, Ceres is commonly used in tightly-coupled CT-LOAM systems.
In our opinion, this is because GTSAM and g2o were originally developed for visual-odometry problems with elaborate factor graphs. However, these factors are, and should be, fixed for the whole lifespan of the graph, and can be used for both short-term (odometry) and long-term (bundle adjustment) optimization processes. On the other hand, LOAM problems have a much simpler factor graph focusing on a short sliding window, and when the map is updated, the plane / edge coefficients in each factor will also change. Hence the cost function has to be rebuilt each time the map is updated.

{SLICT2 is inspired by our previous work SLICT \cite{nguyen2023slict}, where our primary goal is to upgrade the continuous-time trajectory formulation from PLI to B-spline for the benefits mentioned earlier, while still retaining the competitive performance thanks to the use of multi-resolution incremental mapping and feature association scheme. Following the conventional wisdom, we originally employ the NLS in the B-spline optimization process. However the drawback of non-real-time performance in SLICT still persists. Indeed, the real time issue is not exclusive to SLICT but most CT-LOAM methods.} 
By a thorough investigation, it came to our attention that the NLS incurs significant time overheads when building the least-square problem. Moreover, internal procedures to guarantee monotonic decrease of the cost function also requires extra time, rendering the NLS unappealing.
In seeking a remedy for the real time issue, we find that a straightforward approach turns out to yield much better result. Specifically, by forgoing the NLS approach and directly implementing an on-manifold optimization solver based on the Eigen library
\footnote{https://eigen.tuxfamily.org}
(a library for linear algebra operations ubiquitous in robotics research), the CT-LIO problem can converge within a few iterations. In fact, after the first iteration, the state estimate is already good enough for a better association. Therefore, we combine these processes into a solve-associate-solve loop which can converge in 2 or 3 iterations. Effectively, our CT-LIO system acheives both time efficiency (experiments show that it can be up to 8 times faster than using NLS) and competitive accuracy. Moreover, there is no compromise in performance even with a very dense number of control points and lidar factors, which we find to be an issue in other works.
In summary, the contribution of SLICT2 can be stated as follows:

\begin{itemize}
    \item An efficient real-time CT-LIO pipeline with few iterations based on the linear solver of Eigen library and internal feature-to-map association applied between each iteration.
    \item A detailed local-to-global formulation of the residual and Jacobian of the lidar and IMU factors over the B-spline control points, crucial for the computational efficiency of SLICT2.
    \item Extensive experiments and analysis to demonstrate the competitive performance of SLICT2 over state-of-the-art (SOTA) CT-LIO methods, with an in-depth investigation into the time expenditures of the sub-processes.
    \item We release our work for the benefits of the community.
\end{itemize}

The remaining of this paper is organized as follows: in Sec. \ref{sec: preliminary}, we present the basic notations and the theoretical background of the system. Sec. \ref{sec: methodology} provides descriptions of the main steps in the SLICT2 pipeline, and the calculations of the residual and Jacobian of the lidar and inertial factors. In Sec. \ref{sec: experiment}, we perform comparisons of SLICT2 with other SOTA continuous-time lidar odometry methods, and then go deeper into our algorithm to investigate the time cost of each underlying process. Sec. \ref{sec: conclusion} concludes our work.

\section{Preliminaries} \label{sec: preliminary}

\subsection{Notations}
For a quantity $\x$, we use the breve notation $\breve{\x}$ to indicate a measurement of $\x$, and $\hat{\x}$ its estimate. For a matrix $\x$, $\x^\top$ denotes its transpose.
For $\res \in \R^n$ and $\vbf{M} \in \R^{n\times n}$, we denote $\norm{\res}_\vbf{M}^2 \triangleq \res^\top \vbf{M} \res$. For a transformation matrix $\tf \in \SE$, we also use the expression $\tf=(\rot, \pos)$, where $\rot \in \SO$ and $\pos \in \R^3$ are respectively the rotation and translation components of $\tf$.

We denote the \textit{Lie algebra} group (group of skew-symmetric matrices) as $\mathfrak{so}(3)$, and use $(\cdot)_\times$ and $(\cdot)_\vee$ to denote the mappings $(\cdot)_\times \colon \R^3 \to \mathfrak{so}(3)$ and $(\cdot)_\vee \colon \mathfrak{so}(3) \to \R^3$. In addition, the mapping $\Exp : \R^3 \to \SO$ and its inverse $\Log : \SO \to \R^3$ will also be invoked in subsequent parts. We also use the operator $\boxplus : \mathcal{M} \times \R^n \to \mathcal{N} $ to indicate the generalized addition operation on manifolds.
We also recall \textit{the right Jacobian} and \textit{the inverse right Jacobian} of $\SO$ as $J_r$ and $J_r^{-1}$ defined in \cite{sola2018micro, nguyen2021viral}.
Their closed-form formula is as follows:
\begin{align*}
    &J_r(\eulvec) = \ident - \frac{1 - \cos(\norm{\eulvec})}{\norm{\eulvec}^2}\eulvec_\times + \frac{\norm{\eulvec} - \sin(\norm{\eulvec})}{\norm{\eulvec}^3}(\eulvec_\times)^2,
    \\
    &J_r^{-1}(\eulvec) = \ident + \frac{1}{2}\eulvec_\times  +\ll(\frac{1}{\norm{\eulvec}^2} - \frac{1 + \cos(\norm{\eulvec})}{2\norm{\eulvec}\sin(\norm{\eulvec})}\rr)(\eulvec_\times)^2,
\end{align*}
{for $\phi \in \R^3$}.

\subsection{B-spline based interpolation} \label{sec: B-spline basics}

\def\tmin{t_\text{min}}
\def\tmax{t_\text{max}}

A \textit{B-spline} is a piece-wise polynomial defined by the following parameters: \textit{knot length} $\Dt$, spline \textit{order} $N$ (or alternatively the \textit{degree} $D \triangleq N-1$), and the set of \textit{control points} $\ll\{\tf_m\rr\}_{m=0}^{M}$, $\tf_m \in \SE$.
Each control point $\tf_m$ is associated with a \textit{knot}, denoted as $t_m$, where $t_m = t_0 + m\Dt$.
Given $N$, a \textit{blending matrix} $\vbf{B}^{(N)}$ and its \textit{cumulative form} $\tilde{\vbf{B}}^{(N)}$ can be calculated as follows:
\begin{align}
    &\vbf{B}^{(N)} = [b_{m,n}] \in \R^{N\times N};\ m, n \in \{0,\dots N-1\}, \\
    &\tilde{\vbf{B}}^{(N)} = [\tilde{b}_{m,n}];\ \tilde{b}_{m,n} = \sum_{j=m}^{D}b_{j,n}, \label{eq: blending matrix cumulative}\\
    &b_{m,n} = \frac{1}{n!(D-n)!} \sum_{l=m}^{D}(-1)^{l-m} C_{N}^{l-m}(D-l)^{D-n}. \label{eq: blending matrix}
\end{align}
note that $C_{D}^{n} = \frac{D!}{n!(D-n)!}$ denotes the binomial coefficients.

From the control points and the set of knots, for any time $t \in \ll[t_0, t_{M - N + 2}\rr]$, which is the sampling time of some measurement, we can interpolate the pose $\tf(t) = \ll(\rot(t), \pos(t)\rr)$ at this time by the procedure:
\begin{align}
    &s = (t - t_i)/\Dt,\ t \in [t_i. t_{i+1}),\ t_i \in \ll\{t_m\rr\}_{m=0}^M, \label{eq: interval, scaling factor}
    \\
    &[\lambda_0\ \lambda_1\ \dots\ \lambda_{N-1}]^\top = \vbf{B}^{(N)}[1\ s\ \dots\ s^{N-1}]^\top, \label{eq: scaling coefs}
    \\
    &[\tilde{\lambda}_0\ \tilde{\lambda}_1\ \dots\ \tilde{\lambda}_{N-1}]^\top = \tilde{\vbf{B}}^{(N)} [1\ s\ \dots\ s^{N-1}]^\top, \label{eq: scaling coefs cumulative}
    \\
    &\rot(t) = \rot_i\prod_{j=1}^{N-1} \Exp\ll(\tilde{\lambda}_j \Log\ll(\rot_{i+j-1}^{-1}\rot_{i+j}\rr)\rr). \label{eq: interpolated rot}
    \\
    &\pos(t) = \pos_i + \sum_{j=1}^{N-1} \lambda_j \pos_{i+j}. \label{eq: interpolated pos}
\end{align}
In other words, given the time $t$, we first find the knot $t_i$ and the corresponding interval $[t_i, t_{i+1})$ that contains $t$, then calculate the normalized time $s$, as conveyed in step \eqref{eq: interval, scaling factor}. Next, we calculate the scaling coefficients $\lambda_i$, $\tilde{\lambda}_i$, $i = 0, \dots N-1$ by using the blending matrix and the powers of $s$ in steps \eqref{eq: scaling coefs} and \eqref{eq: scaling coefs cumulative}. Finally, we calculate the interpolated pose $\ll(\rot(t), \pos(t)\rr)$ at time $t$ by the scaling coefficients and $N$ control points at knots $t_i, \dots t_{i+N-1}$ in \eqref{eq: interpolated rot} and \eqref{eq: interpolated pos}.

\begin{remark}
Note that for a predefined spline, the above procedure is only applicable for sampling time $t$ not exceeding $t_{M - N + 2}$.
Inversely, to represent a trajectory over an interval $[\tmin, \tmax]$ by a B-spline with knot length $\Dt$, we set $t_0 = \tmin$ and the last knot will have the index $M = \ceil{(\tmax - \tmin)/\Dt} + N - 2$.
In Fig. \ref{fig: discrete vs continuous}, a sliding window with $M=4$ and $N=3$ is shown. Since $N=3$, one control point at time $t_{k+1}$ has to be added for the B-spline to be completely defined on the interval $[t_w, t_k]$, where all of the measurements are being considered.
The basalt\footnote{\url{https://vision.in.tum.de/research/vslam/basalt}} library \cite{sommer2020efficient} is used to aid in the B-spline formulation in our work.
\end{remark}

For convenience in later calculations of residual and Jacobian, we also define the following quantities:
\begin{align}
    &[\dot{\tilde{\lambda}}_0\ \dot{\tilde{\lambda}}_1\ \dots\ \dot{\tilde{\lambda}}_{N-1}]^\top = \tilde{\vbf{B}}^{(N)}[ 0\ 1\ \dots\ \dot{s}^{j} \dots\ \dot{s}^{N-2}]^\top,
    \\
    &[ \ddot{{\lambda}}_0\ \ddot{{\lambda}}_1\ \dots\ \ddot{{\lambda}}_{N-1}]^\top = \vbf{B}^{(N)}[ 0\ 0\ \dots\ \ddot{s}^{j} \dots\ \ddot{s}^{N-2} ]^\top, \label{eq: scaling factor ddot}
    \\
    &\dbf_j = \Log\ll(\rot_{i+j-1}^{-1}\rot_{i+j}\rr),\quad \Abf_j = \Exp\ll(\tilde{\lambda}_i \dbf_j\rr),
    \\
    &\Pbf_{j-1} = \Pbf_j\Abf_j^\top,\quad \Pbf_{N-1} = I,
    \\ 
    &\angvel^{(j)}
    =
    \Abf^\top_{j-1}\angvel^{(j-1)} + \dot{\tilde{\lambda}}_{j-1}\dbf_{j-1}, \label{eq: angvel calculation 2}
    \ 
    \angvel^{0} = \angvel^{1} = \vbf{0},
    \\
    &\angvel_t = \angvel^{(N)}. \label{eq: angvel calculation}
\end{align}
and note that $\dot{s}^j = \frac{j}{\Dt}s^{j-1}$ and $\ddot{s}^j = \frac{j(j-1)}{\Dt^2}s^{j-2}$.

\subsection{Observation models}

Most sensing modality can be characterized by an observation model as follows:
\begin{align} \label{eq: observation model}
    h(\X, \Z, \M, \eta) = 0,
\end{align}
where $\X$ is the state to be estimated, $\Z$ is some measurement from the sensor, $\M$ is the prior information, and $\eta$ denotes the noise in the measurement or error in modeling.

If we denote $\hat{\X}$ as an estimate of $\X$, the quantity $h(\hat{\X}, \Z, \M, 0)$ can be referred to as the residual. In some cases $\M$ can be absent or considered part of the observation model $h(\cdot)$. 
Thus, we may write $h(\X, \Z, \eta) = 0$ and $h(\hat{\X}, \Z, 0)$ as the observation model and the residual, respectively.

In the case of CT-LIO, the IMU and lidar observation models can be expressed as follows:
\begin{align}
    &(\rot_t^{-1}\dot{\rot}_t)_\vee + \bias_\omega + \vbf{\eta}_\omega - \breve{\angvel}_t = 0, \label{eq: omega}
    \\
    &\rot_t^{-1}(\Ddot{\pos}_t + \grav) + \bias_a + \vbf{\eta}_a - \breve{\accel}_t = 0 \label{eq: accel}
    \\
    &\n^\top (\rot_t {}^{\fB_t}\breve{\f} + \pos_t) + \mu - \eta_l = 0, \label{eq: point to plane}
\end{align}
where $\rot$ and $\pos$ are the orientation and position states; $\breve{\angvel}_t$, $\breve{\accel}_t$, ${}^{\fB_t}\breve{\f}$ are the angular velocity, acceleration, and lidar range measurements at the time instant $t$; $\bias_\omega$ and $\bias_a$ are the IMU biases; $\vbf{\eta}_\omega$, $\vbf{\eta}_a$, $\eta_l$ are the measurement noise; $\grav$ is the gravity constant; and $\n$, $\mu$ are the plane coefficients of a neighborhood in the map associated with $\breve{\f}$.

To simplify the notation, in the subsequent parts we will denote $r_\L \triangleq h(\hat{\X}, {}^{\fB_t}\f, 0)$ and $r_\I \triangleq h(\hat{\X}, (\angvel_t. \accel_t), 0)$ as the lidar and IMU residuals.

\subsection{The state estimate}

The trajectory of the robots represented by a spline starting from the start time $t_0$, and extended as new data are acquired. To keep the problem bounded, we only optimize the control points corresponding to the sliding window spanning the latest interval $[t_s, t_e]$. Specifically, the states to be estimated are:
\begin{equation} \label{eq: state estimate}
    \X = \ll(\tf_w, \tf_{w+1}, \dots \tf_{k+N-2}, \bias_\omega, \bias_a\rr),
\end{equation}
where $\tf_w, \tf_{w+1}, \dots \tf_{k+N-2}$ are the control points on the sliding window, with $[t_s, t_e] \subset [t_w, t_{k+N-2}]$, and $\bias_\o, \bias_a$ are the IMU biases.

Different from discrete-time case where velocity state is often included, in CT-LIO, it is implied in the first order derivative of the position states. Note that one can also include other states such as the local gravity constant, the lidar-IMU offset, and the timestamp offset. However these extra features are beyond the scope of this paper and we believe users can easily modify the basic pipeline of SLICT2 to address these issues should the need arises.

\section{Methodology} \label{sec: methodology}

\subsection{Main Workflow} \label{sec: main workflow}

\begin{figure}
    \centering
    \begin{overpic}[width=0.8\linewidth,
                   ]{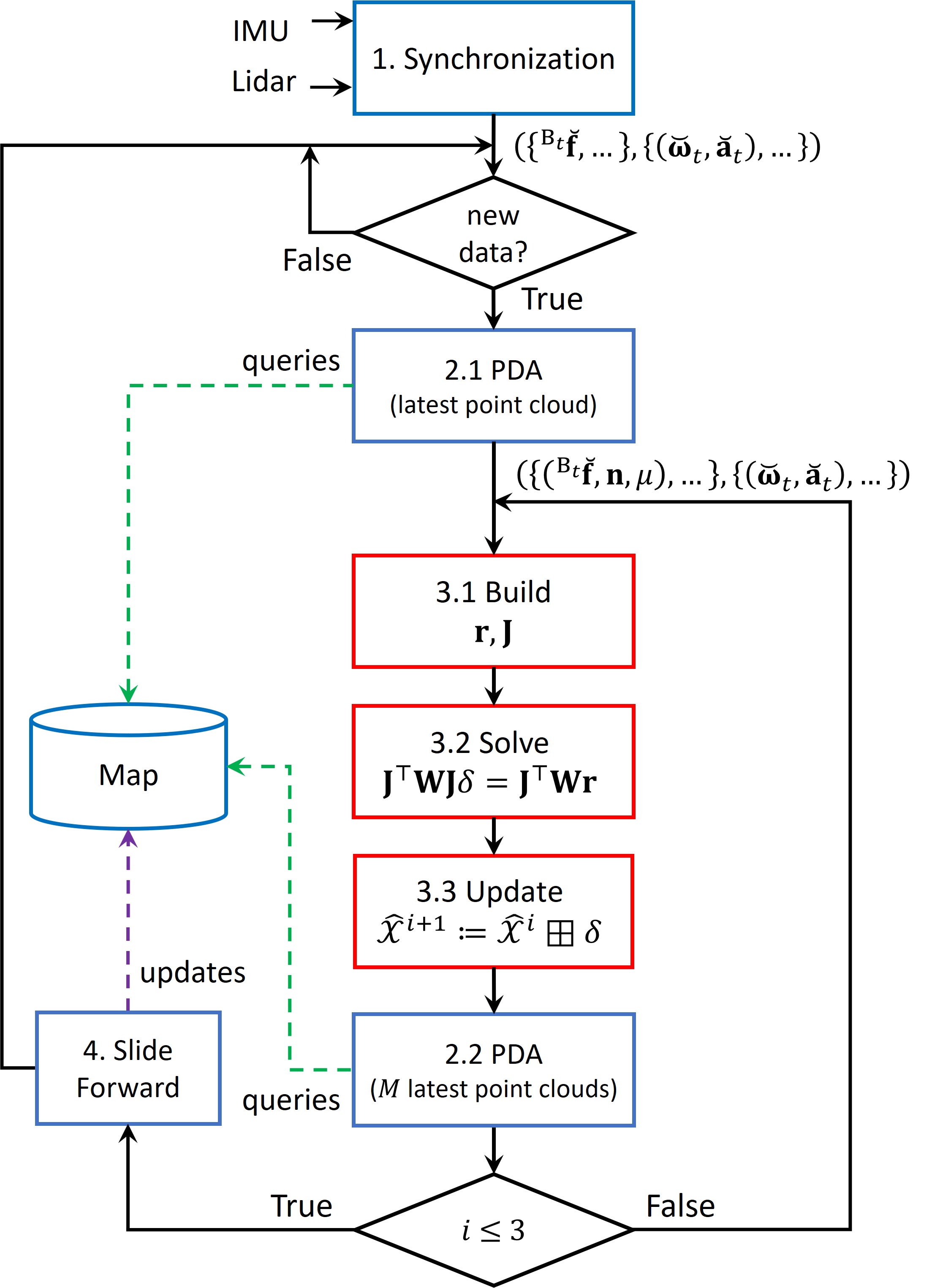}
    \end{overpic}
\caption{SLICT2 is a CT-LIO system with a simple yet efficient pipeline.
$i$ is the index of the inner loop, starting from 0.
We refer to the sequence 3.1 -- 3.2 -- 3.3 -- 2.2 -- 3.1 as the \textit{inner loop}, and the \textit{outer loop} refers to the sequence 2.1 -- (inner loops) --- 2.1.
Description of each block is given in Sec. \ref{sec: main workflow}.
}
\label{fig: main pipeline}
\vspace{-0.25cm}
\end{figure}

{Fig. \ref{fig: main pipeline} presents the main workflow of SLICT2. The numbering of these blocks conincide with the sections below.
The blocks in blue (1, 2.1, 2.2, 4) are inspired by SLICT, which are briefly recapped to provide some background information. The original contributions of SLICT2 are highlighted in red (3.1, 3.2, 3.3), which will be elaborated in details.}

\subsubsection{Synchronization}
In the synchronization process, the IMU and lidar data are buffered and organized in so-called bundles. Each bundle consists of a set of raw lidar points $\{{}^{\fB_t}\breve{\f},\dots\}$ and a set of IMU measurements $\{(\breve{\angvel}_t, \breve{\accel}_t),\dots\}$. We also carefully keep track of the timestamps to maintain the time limits of the sliding window.

\subsubsection{Propagation-Deskew-Association}

Most LIO methods employ the point-to-plane observation model $\n \f + \mu = 0$. To obtain the plane coefficients $(\n, \mu)$ for each feature $\breve{\f}$, one needs to conduct the following so-called PDA (propagation-deskew-association) procedure:

\tb{IMU propagation}: When a new bundle of measurements $\ll(\{{}^{\fB_t}\breve{\f},\dots\}, \{(\breve{\angvel}_t, \breve{\accel}_t),\dots\}\rr)$ is obtained, we will use the IMU measurements $\{(\breve{\angvel}_t, \breve{\accel}_t),\dots\}$ to propagate the states and obtain the poses at each IMU sampling time within the step time. This process relies on an initial guess of the state at the first sample.
This is why we are motivated to redo propagation after each iteration as we can achieve a better state estimate.

\tb{Deskew (motion undistortion)}: In this step, the \textit{world}-referenced coordinate of ${}^{\fB_t}\breve{\f}$, denoted as ${}^\fW\breve{\f}$, is obtained by the transform 
${}^\fW\breve{\f} = \breve{\rot}_t {}^{\fB_t}\breve{\f} + \breve{\pos}_t,$
where $(\breve{\rot}_t, \breve{\pos}_t)$ is the pose estimate at the sampling time $t$ of the lidar point $\breve{\f}$. This pose estimate can be obtained from sampling the spline (for the old bundles), or by using IMU-propagated pose (for the latest bundle). Again, due to inaccuracy in the propagation step, the deskew process may not be perfect. Hence, a redo can be beneficial after a better state estimate is achieved.

\tb{Association}: The value ${}^\fW\breve{\f}$ is used to find a neighbourhood in the map (which can be done by kNN search \cite{nguyen2021miliom, xu2021fast}, or proximity to a surfel \cite{park2018elastic, nguyen2023slict}). Based on the distribution of points in the neighbourhood plane coefficients $(\n, \mu)$ can be calculated and associated with $\breve{\f}$ for the construction of the point-to-plane factor.
Since the two steps need to be done together, we shall refer to both steps above as the \textit{association process}.

\begin{remark}
Apparently,
all steps in the PDA process rely on some initial estimate of the trajectory, which may not be accurate before thorough optimization.
One approach to resolve this issue is to add all possible associable planes \cite{koide2021voxelized}. However this approach can be computationally demanding, and wrong associations still have an impact on the overall solution. Therefore, in SLICT2, we opt to simply redoing the association after every update. This is found to be feasible when update step can be done more efficiently with a simple approach, leaving sufficient time for redoing the PDA process after each update.
\end{remark}

\subsubsection{Build, Solve and Update}

When all of the measurements $\Z$ on the sliding window have been accounted for, we can begin the so-called Build-Solve-Update (BSU) process. In essence, this process seeks to minimize a cost function $f$ by tuning the state estimate $\hat{\X}$ as follows:

\begin{align}
    f
    &= \sum \norm{r_\L(\hat{\X}, {}^{\fB_t}\breve{\f})}_{\vbf{W}_\L}^2
    \nonumber
    \\
    &\qquad+ \sum \norm{r_\I(\hat{\X}, \breve{\angvel}_t, \breve{\accel}_t)}_{\vbf{W}_\I}^2
    \triangleq
    \norm{h(\hat{\X}, \Z)}^2_{\vbf{W}}, \label{eq: map problem}
\end{align}
where $\res_\I$ and $\res_\L$ are the IMU and lidar resdiuals, $\vbf{W}_\I$, $\vbf{W}_\L$, $\vbf{W}$ are some weight matrices (often the inverse of covariance), and $h(\hat{\X}, \Z)$ is the combined residual vector at the global scope.

To minimize \eqref{eq: map problem}, we start with an initial guess $\hat{\X}^0$, then iteratively update $\hat{\X}^i$ by a small perturbation $\d^* \triangleq \argmin_{\d} f(\hat{\X}^{i}\boxplus\d), i = 1, 2, \dots$ for multiple steps.
To solve the problem $\d^* \triangleq \argmin_{\d} f(\hat{\X}^{i}\boxplus\d)$ in each iteration, we apply linearization at $\hat{\X}^i$ as follows:
\begin{align} \label{eq: map linearized}
    f(\hat{\X}^{i}\boxplus\d)
    &= 
    \norm{h(\hat{\X}^{i}\boxplus\d,\Z)}^2_{\vbf{W}}
    \approx \norm{\res + \Jcb\d}^2_{\vbf{W}}
    \nonumber
    \\
    &=
    \d^\top \Jcb^\top \vbf{W} \Jcb \d + 2\d^\top\Jcb^\top \vbf{W} \res + \res^\top \vbf{W} \res,
\end{align}
where $\res = h(\hat{\X}^{i},\Z)$, and $\Jcb$ is the Jacobian $\frac{\partial h(\hat{\X}^i\boxplus\d,\Z)}{\partial \d}\Big|_{\d=\vbf{0}}$. The details of $\res$ and $\Jcb$ are discussed in Sec.\ref{sec: residual and jacobian}.
The minimum of \eqref{eq: map linearized} is the solution of the linear system:
\begin{equation} \label{eq: solution}
    \Jcb^\top \vbf{W} \Jcb\d = - \Jcb^\top\vbf{W} \res.
\end{equation}
which is of the form $AX=B$, and can be solved by a variety of techniques.

\begin{remark}
As illustrated in Fig. \ref{fig: rj structure}, each inner loop consists of one run of the steps 3.1, 3.2, 3.3, 2.2.
In NLS-based methods, the steps 3.1, 3.2, 3.3 are handled by a solver, and they can be iterated inside the solver.
Empirically, we notice that Ceres would need 7 or 8 iterations for convergence, whereas in SLICT2, we find that we only need 2 or 3 iterations to achieve convergence.
\end{remark}

\subsubsection{Slide Window}

At the end of each outer loop, we will judge if the oldest data on the sliding window can be marginalized as a key frame. If the smallest translational or rotational distance of this key frame candidate from $K$ nearest key frames is larger than a threshold, the candidate can be made into a new key frame. Its point cloud is then inserted to a global map that is used for the PDA process. Our implementation supports both ikdTree \cite{xu2021fast} and surfel tree \cite{nguyen2023slict}. Empirically we find the surfel tree offers better stability and accuracy than ikd-Tree.

\subsection{Residual and Jacobian for B-spline-based factors} \label{sec: residual and jacobian}

In this section we provide the detailed calculation of the residual and the analytical Jacobian of the lidar and inertial factors, which are essential for ensuring real-time performance of the method.

\subsubsection{Lidar factor}

For each lidar feature ${}^{\fB_t}\breve{\f}$,
based on the observation model \eqref{eq: point to plane} and the interpolation rules in Sec. \ref{sec: B-spline basics},
the residual of the lidar factor can be expressed explicitly as a function of the state estimate (the control points) and the measurements can be put as follows:
\begin{align} \label{eq: lidar residual}
    &r_\L
    = \n^\top
    \Bigg[\hat{\rot}_i\prod_{j=1}^{N-1} \Exp\ll(\tilde{\lambda}_i \Log\ll(\hat{\rot}_{i+j-1}^{-1}\hat{\rot}_{i+j}\rr)\rr){}^{\fB_t}\breve{\f}
    \nonumber
    \\
    &\qquad\qquad\qquad + \hat{\pos}_i + \sum_{j=1}^{N-1} \lambda_i \ll(\hat{\pos}_{i+j} - \hat{\pos}_{i+j-1}\rr)
    \Bigg] + \mu,
    \nonumber
\end{align}
where $t \in [t_i, t_{i+1})$.
However to handle it more easily, we can write \eqref{eq: lidar residual} as:
\begin{equation} \label{eq: lidar residual}
    r_\L = \n^\top(\hat{\rot}_t{}^{\fB_t}\breve{\f} + \hat{\pos}_t) + \mu,
\end{equation}
where $(\hat{\rot}_t, \hat{\pos}_t)$ is an auxiliary state estimate that is a combination of the decision variables (the control points) that are coupled with $\breve{\f}$.
Apparently each factor $r_\L$ is coupled with $N$ control points, hence its Jacobian over the state estimate $\hat{\X}$ has the form:
\begin{align}
    &\frac{\partial r_\L}{\partial \hat{\X}}
    =
    \begin{bmatrix}
    \dots &\dfrac{\partial r_\L}{\partial \hat{\rot}_{i+j}} &\dfrac{\partial r_\L}{\partial \hat{\pos}_{i+j}} &\dots
    \end{bmatrix},
\end{align}
for $i \in [w, k+N-2]$ such that $[t_i, t_i+1) \ni t, t_i \in \{t_s, \dots t_{k+N-2}\}, j \in \{0, N-1\}$.
An illustration of this Jacobian is provided in Fig. \ref{fig: rj structure}.
To calculate the Jacobian of $r_\L$ over $\hat{\rot}_{i+j}$ and $\hat{\pos}_{i+j}$, $j \in \{0, 1, \dots, N-1\}$, we apply the chain rule as follows:
\begin{align}
    &\frac{\partial r_\L}{\partial \hat{\rot}_{i+j}}
    =
    \frac{\partial r_\L}{\partial \hat{\rot}_t}
    \frac{\partial \hat{\rot}_t}{\partial \hat{\rot}_{i+j}},
    \quad 
    \frac{\partial r_\L}{\partial \hat{\pos}_{i+j}}
    =
    \frac{\partial r_\L}{\partial \hat{\pos}_t}
    \frac{\partial \hat{\pos}_t}{\partial \hat{\pos}_{i+j}}.
    \nonumber
\end{align}
Then the task now is to calculate the Jacobians $\dfrac{\partial r_\L}{\partial \hat{\rot}_t}$, $\dfrac{\partial r_\L}{\partial \hat{\pos}_t}$, $\dfrac{\partial \hat{\rot}_t}{\partial \hat{\rot}_m}$, and $\dfrac{\partial \hat{\pos}_t}{\partial \hat{\pos}_m}$. The first two can be easily calculated as:
\begin{align}
    \frac{\partial r_\L}{\partial \hat{\rot}_t} = \n^\top\hat{\rot}_t{}^{\fB_t}\breve{\f}_\times,
    \qquad
    \frac{\partial r_\L}{\partial \hat{\pos}_t} = \n^\top,
\end{align}
whereas $\dfrac{\partial \hat{\rot}_t}{\partial \hat{\rot}_m}$ and $\dfrac{\partial \hat{\pos}_t}{\partial \hat{\pos}_m}$ can be found by:
\begin{align}
    &\frac{\partial \rot_t}{\partial \rot_{i+j}}
    =
    \tilde{\lambda}_j
    \Pbf_j
    J_r(\tilde{\lambda}_j\dbf_j)J_r^{-1}(\dbf_j)
    \nonumber
    \\
    &\qquad\qquad
    -
    \tilde{\lambda}_{j+1}
    \Pbf_{j+1}
    J_r(\tilde{\lambda}_{j+1}\dbf_{j+1})J_r^{-1}(-\dbf_{j+1}),
    \label{eq: spline jacobian rot}
    \\
    &\frac{\partial \pos_t}{\partial \pos_{i+j}}
    =
    \lambda_j I. \label{eq: spline jacobian pos}
\end{align}
Equations \eqref{eq: spline jacobian rot} and \eqref{eq: spline jacobian pos} are revised from the findings of Sommer \textit{et al} \cite{sommer2020efficient} in a more concise and self-contained formula. Note that for the boundary case of $j=0$, we maintain that \eqref{eq: spline jacobian rot} can still be applied by defining $\Abf_0 = \rot_i$ and $\dbf_0 = \Log(\rot_i)$; whereas when $j = N-1$, we define $\tilde{\lambda}_{N} = 0$ and thus one can zero out the second term in \eqref{eq: spline jacobian rot}.

\subsubsection{IMU factor}
The residual of IMU factors is defined as:
\begin{equation}
    r_\I
    =
    \begin{bmatrix}
    r_\o
    \\
    r_a
    \\
    r_{b_\o}
    \\
    r_{b_a}
    \end{bmatrix}
    =
    \begin{bmatrix}
    \hat{\angvel}_t
    + \hat{\bias}_\omega - \breve{\angvel}_t
    \\
    \hat{\accel}_t
    + \hat{\bias}_a - \breve{\accel}_t
    \\
    \hat{\bias}_\o - \bar{\bias}_\o
    \\
    \hat{\bias}_a - \bar{\bias}_a
    \end{bmatrix} \in \R^{12},
\end{equation}
where $\hat{\angvel}_t$ and $\hat{\accel}_t$ are the predicted angular velocity and body-referenced acceleration, $\bar{\bias}_\o$ and $\bar{\bias}_a$ are some prior values of the biases, which are the same as the estimates $\hat{\bias}_\o$ and $\hat{\bias}_a$ at the end of the previous outer loop.
Note that $\hat{\angvel}_t$ can be calculated by the formulas \eqref{eq: angvel calculation} and \eqref{eq: angvel calculation 2}, and $\hat{\accel}_t$ from \eqref{eq: scaling factor ddot} and \eqref{eq: interpolated pos}.

We see that $r_\I$ is coupled with $N$ control points and the two bias states. Thus its Jacobian over the state estimate of \eqref{eq: state estimate} will have the following form:
\begin{align} \label{eq: lidar jacobian}
    &\frac{\partial r_\I}{\partial \hat{\X}}
    =
    \begin{bmatrix}
    \dots &\dfrac{\partial r_\o}{\partial \hat{\rot}_{i+j}} &0                                              &\dots &I &0\\
    \dots &\dfrac{\partial r_a}{\partial \hat{\rot}_{i+j}}  &\dfrac{\partial r_a}{\partial \hat{\pos}_{i+j}} &\dots &0 &I\\
    \dots &0                                               &0                                              &\dots &I &0\\
    \dots &0                                               &0                                              &\dots &0 &I\\
    \end{bmatrix}
    ,
\end{align}
Fig. \ref{fig: rj structure} provides an illustration of the structure of this Jacobian.
Now, we only need to compute the following Jacobians to complete $\dfrac{\partial r_\I}{\partial \hat{\X}}$:
\begin{equation*}
    \frac{\partial r_\o}{\partial \hat{\rot}_{i+j}} = \frac{\partial \hat{\angvel}_t}{\partial \hat{\rot}_{i+j}},
    \ 
    \frac{\partial r_a}{\partial \hat{\rot}_{i+j}} = \frac{\partial \hat{\accel}_t}{\partial \hat{\rot}_{i+j}},
    \ 
    \frac{\partial r_a}{\partial \hat{\pos}_{i+j}} = \frac{\partial \hat{\accel}_t}{\partial \hat{\pos}_{i+j}}.
\end{equation*}

Indeed, the later two can be easily calculated as follows:
\begin{equation}
    \frac{\partial \hat{\accel}_t}{\partial \hat{\rot}_{i+j}} = (\hat{\accel}_t)_\times\frac{\partial \hat{\rot}_t}{\partial \hat{\rot}_{i+j}},
    \quad
    \frac{\partial \hat{\accel}_t}{\partial \hat{\pos}_{i+j}} = \hat{\rot}_t^{-1}\ddot{\lambda_j},
\end{equation}
where $\dfrac{\partial \hat{\rot}_t}{\partial \hat{\rot}_{i+j}}$ follows \eqref{eq: spline jacobian rot}, and $\ddot{\lambda}_j$ follows \eqref{eq: scaling factor ddot}.

Finally, for the Jacobian $\dfrac{\partial \hat{\angvel}_t}{\partial \hat{\rot}_{i+j}}$, it can be calculated via the following recursive equation:
\begin{align}
    &\frac{\partial {\angvel}^{(j)}}{\partial {\dbf}_{j}}
    =
    \Pbf_j\ll(\tilde{\lambda}_j\Abf^\top_j\angvel^{(j)}_\times J_r(-\tilde{\lambda}_j\dbf_j) + \dot{\tilde{\lambda}}_jI\rr).
    \\
    &\frac{\partial {\angvel}_t}{\partial {\rot}_{i+j}}
    =
    \frac{\partial {\angvel}^{(j)}}{\partial {\dbf}_{j}}J_r^{-1}(\dbf_j)
    - 
    \frac{\partial {\angvel}^{(j+1)}}{\partial {\dbf}_{j+1}}J_r^{-1}(-\dbf_{j+1}), \label{eq: domega / dR}
\end{align}
In \eqref{eq: domega / dR}, for the boundary case $j=0$ and $j=N-1$, notice that $\dfrac{\partial {\angvel}^{(0)}}{\partial {\dbf}_{0}} = \dfrac{\partial {\angvel}^{(N)}}{\partial {\dbf}_{N}} = \vbf{0}$.

\subsection{Parallelization} \label{sec: parallel}

The calculation of residual and Jacobian in the previous section is just the first step. To form the equation \eqref{eq: solution}, we still need to map these local computations to the global residual and Jacobian matrices. Apparently this is one of the convenience offered by NLS. However, this can be done more efficiently in a parallelization scheme.
First, from the number of IMU and lidar factors, as well as the number of state estimates, we can deduce the appropriate dimensions to allocate the $\res$ and $\Jcb$ matrices. Then, we create concurrent threads to calculate and assign values to the sub-blocks of $\res$ and $\Jcb$ by the appropriate indices.

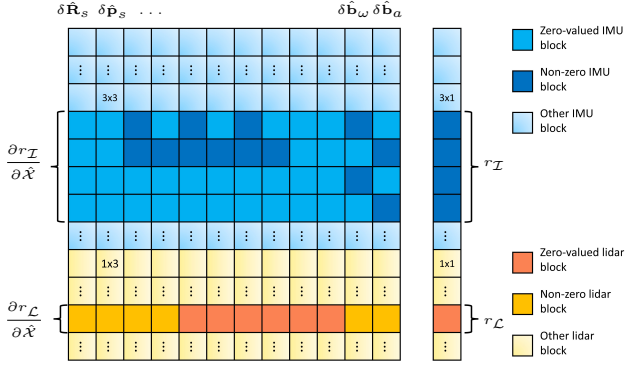
\begin{figure}
    \centering
    \begin{overpic}[width=0.9\linewidth,
                   ]{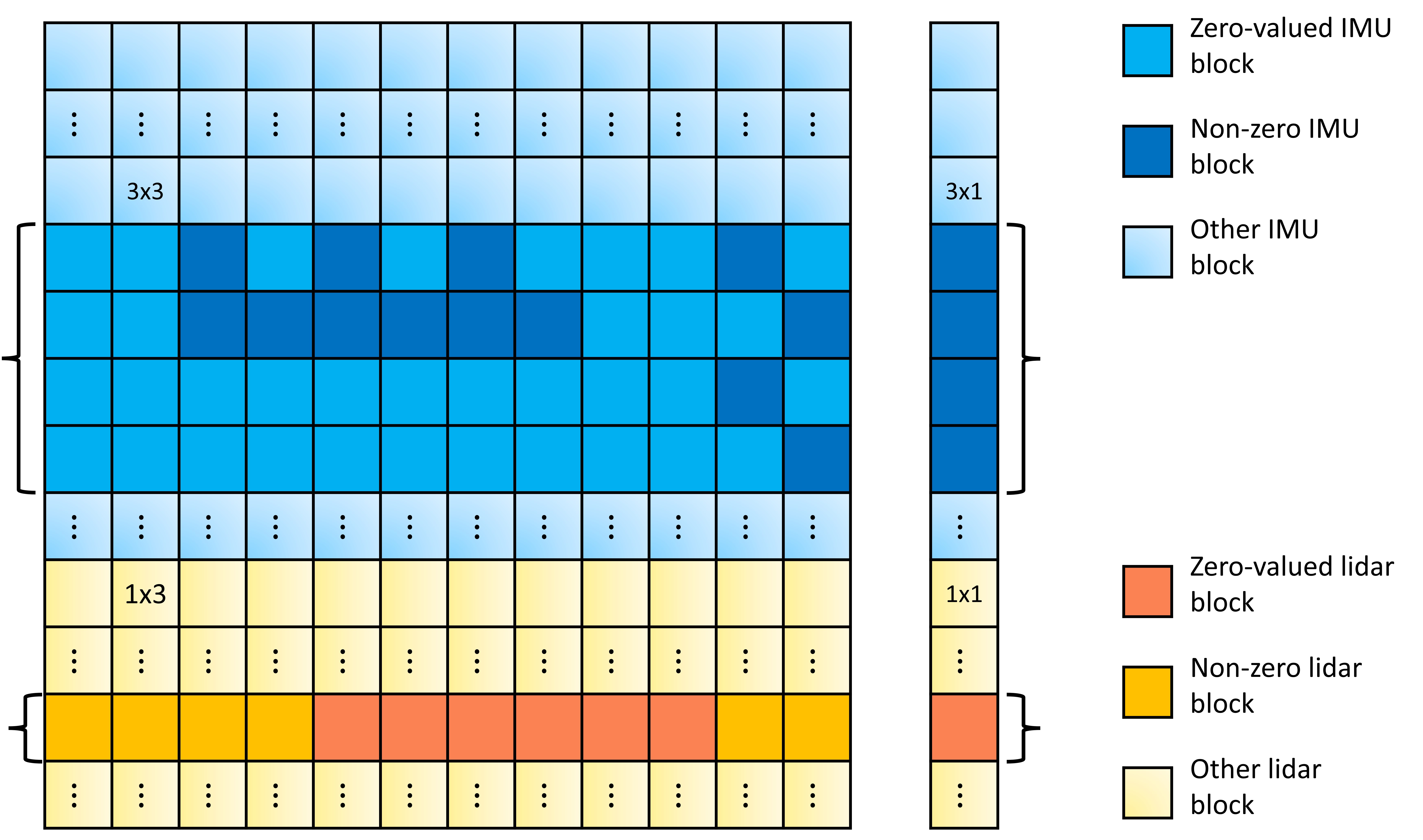}
                   
                   \put(74, 34){\tiny $r_\I$ }
                   \put(-8, 34){\tiny $\dfrac{\partial r_\I}{\partial \hat{\X}}$ }

                   \put(74, 7){\tiny $r_\L$ }
                   \put(-8, 7){\tiny $\dfrac{\partial r_\L}{\partial \hat{\X}}$ }
                    
                   \put(1, 60){\tiny \rotatebox{0}{$\d\hat{\rot}_{s}$}}
                   \put(8, 60){\tiny \rotatebox{0}{$\d\hat{\pos}_{s}$}}
                   \put(15, 60){\tiny \rotatebox{0}{$\dots$}}


                   \put(49, 60){\tiny $\d\hat{\bias}_{\o}$}
                   \put(55, 60){\tiny $\d\hat{\bias}_{a}$}
                    

    \end{overpic}
\caption{The structure of the Jacobian (left) and residual (right). The coupling of state estimates with IMU and lidar measurements in this illustration reflect the graph at Fig. \ref{fig: discrete vs continuous}. Note that the block size represented by each cell is different for IMU and lidar factor. More detailed descriptions are given in Sec. \ref{sec: residual and jacobian} and Sec. \ref{sec: parallel}.
}
\label{fig: rj structure}
\end{figure}

An example is used to illustrate this process in reference to Fig. \ref{fig: rj structure}. Let us denote $K_\I$ and $K_\L$ as the number of IMU and lidar factors, $S=k-w+1+N-2$ as the number of control points (as illustrated in Fig. \ref{fig: discrete vs continuous}), and $X = 6S + 6$ the dimension of all state estimates (consisting of $S$ control points and 2 bias vectors). We can then index the entries in $\Jcb$ and $\res$ as $\Jcb=[J_{m,n}]$ and $\res=[r_{m}]$, where $m \in \{0, \dots K_\I + K_\L -1\}$ and $n \in \{0, \dots X - 1\}$. In our case the residual and Jacobian of IMU factors are stacked above the lidar factors.

Now, for a lidar factor with index $a \in \{0, \dots, K_\L-1\}$, from \eqref{eq: lidar residual} and \eqref{eq: lidar jacobian}, we can find that its residual is the single entry $r_{12.K_\I + a}$ in the vector $\res$, while the non-zero entries of its Jacobian reside at $J_{12.K_\I + a, n}$, $n \in \{6i, \dots 6i + 6N-1\}$, where $i$ is the index of the knot whose interval $[t_i, t_{i+1})$ contains $t$, the sampling time of the lidar measurement.

Similarly, for an IMU factor with index $b \in \{0, \dots, K_\I-1\}$, its residual includes the entries $r_{12b + m}$, $m \in\{0, \dots 12\}$, whereas its non-zero Jacobian spreads across the entries $J_{12b + m, n}$, where $(m, n) \in \{0,\dots 6\} \times \{6i,\dots 6i + 6N-1\} \cup \{0, 12\} \times \{6S,\dots 6S + 6 - 1\}$.

The above analysis is essential to calculating the Jacobian and residuals of thousands of factors in a few miliseconds.
The time efficiency of this process is reported in Sec. \ref{sec: time analysis}.


\section{Experiment} \label{sec: experiment}

\subsection{Accuracy test}

\begin{figure*}
    \centering
    \includegraphics[width=0.45\linewidth]{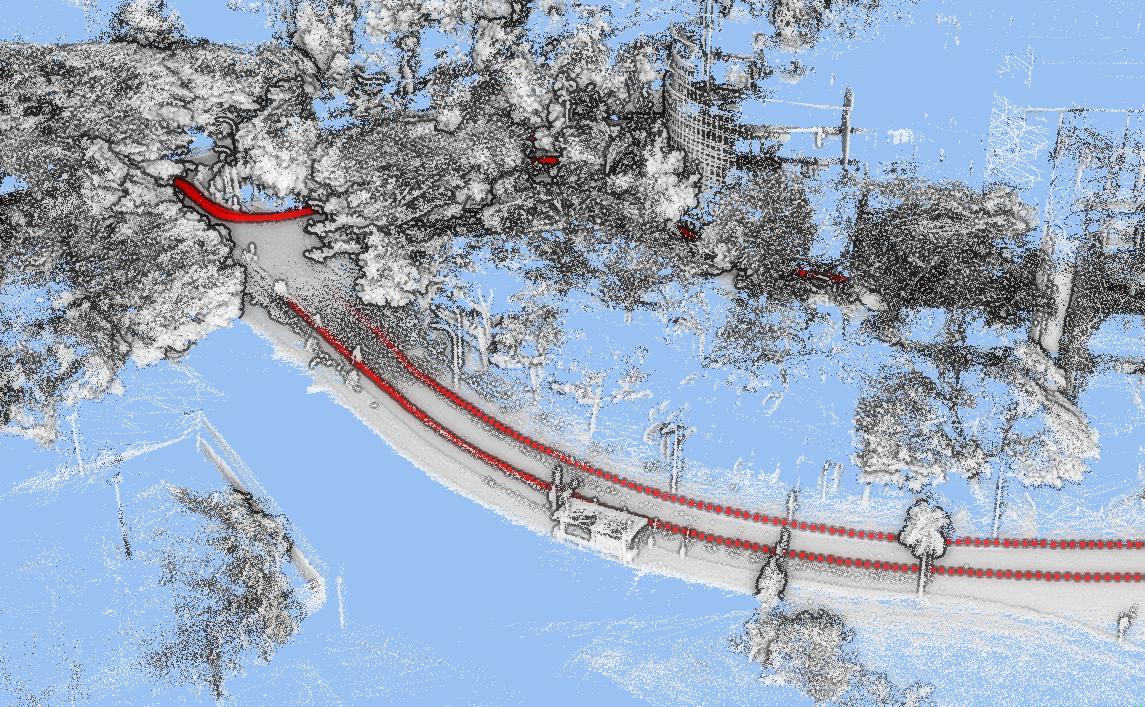}
    \includegraphics[width=0.454\linewidth]{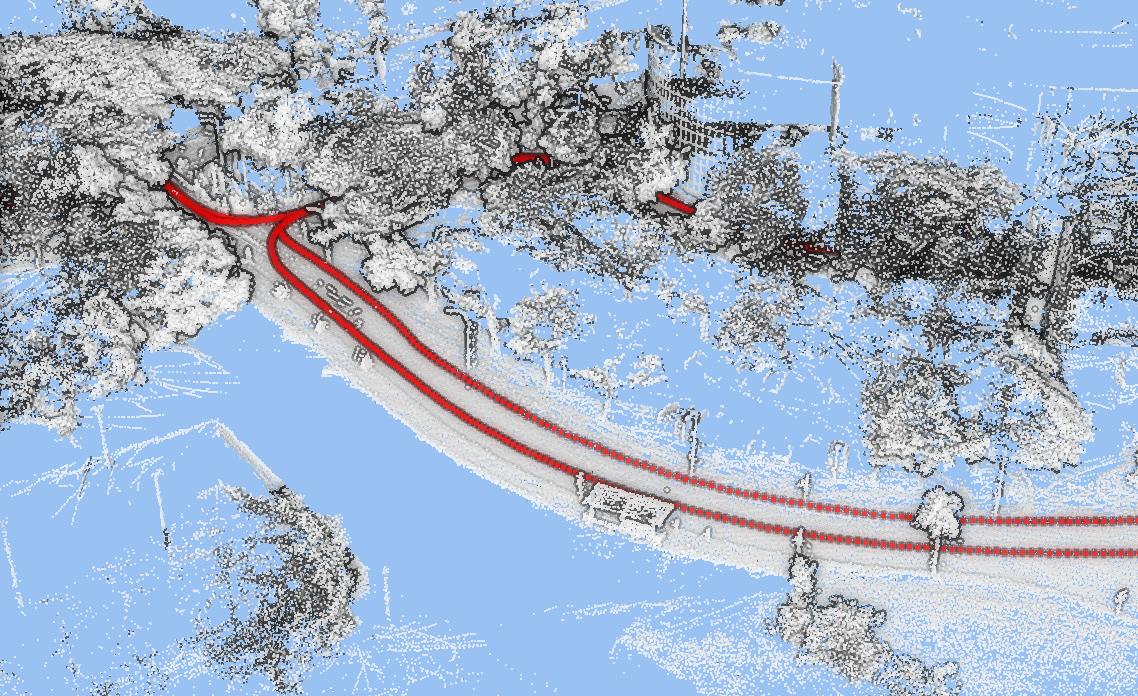}
\caption{The map and trajectory (red) of FAST-LIO (left) and SLICT2 (right) in xxx\_day\_01 sequence. We can see that FAST-LIO has a slight drift in the vertical direction, leading to its larger APE in this sequence.}
\label{fig: maps}
\end{figure*}

In this first part we will compare SLICT2 with other state of the art LIO methods in popular public datasets.


On the methods, the following are selected: FAST-LIO2 \cite{xu2022fast}, CT-ICP\cite{dellenbach2022ct}, SLICT\cite{nguyen2023slict}, and CLIC\cite{lv2023continuous}. FAST-LIO2 is a seminal discrete-time based LIO method that pioneered the iterated Extended Kalman Filter (iEKF) in LIO. It has also inspired many other works \cite{lin2022r3live, bai2022faster, zheng2022fast, chen2024ig} in recent years. For continuous-time, CT-ICP is one of the first publicly released CT-LO methods that use PLI. Similarly, SLICT uses a PLI formulation, but at the sub-interval level of the lidar scan. Finally, CLIC is an improved version of CLINS \cite{lv2021clins}, a LIO method that uses NLS-based B-spline continuous-time formulation.

On the datasets, we select NTU VIRAL \cite{nguyen2021ntuviral} as the first benchmark, in which we use the horizontal 16-channel lidar mounted on an aerial vehicle, the high accuracy high frequency IMU, and a centimeter ground truth captured by a laser tracker.
Since NTU VIRAL sequences have relatively low speed, we select the high velocity sequences in the Multi-Campus Dataset (MCD) \cite{mcdviral2024} as the next benchmark. Moreover, for the MCD benchmark we also experiment with the non-repetitive lidar (livox), which has gained significant popularity in recent years due to low cost and high accuracy, but also poses a challenge with the narrower field of view.

\begin{table}
\centering
\renewcommand{\arraystretch}{1.25}
\begin{threeparttable}

\caption{APE of LO and LIO methods on the NTU dataset.} \label{tab: ntu ate}


\begin{tabular}{|cccccc|}
\hline
\bf Sequence 
          &  \bf  FLIO & \bf CLIC   & \bf CTICP  &  \bf SLICT & \bf SLICT2 \\
\hline\hline
  eee\_01 &      0.069 & 0.056 &  6.503 & \tb{0.024} & \ul{0.025} \\
  eee\_02 &      0.069 & 0.056 &      x & \tb{0.019} & \ul{0.023} \\
  eee\_03 &      0.111 & 0.062 &      x & \ul{0.028} & \tb{0.024} \\
  nya\_01 &      0.053 & 0.069 &  0.459 & \ul{0.022} & \tb{0.021} \\
  nya\_02 &      0.090 & 0.075 &  0.618 & \ul{0.026} & \tb{0.023} \\
  nya\_03 &      0.108 & 0.079 &  0.129 & \ul{0.030} & \tb{0.021} \\
  rtp\_01 &      0.125 & 0.263 &  9.675 & \tb{0.056} & \ul{0.059} \\
  rtp\_02 &      0.131 & 0.150 &  8.907 & \ul{0.062} & \tb{0.056} \\
  rtp\_03 &      0.137 & 0.077 &  9.894 & \ul{0.070} & \tb{0.064} \\
  sbs\_01 &      0.086 & 0.071 &      x & \tb{0.025} & \ul{0.025} \\
  sbs\_02 &      0.078 & 0.073 &  3.872 & \ul{0.033} & \tb{0.032} \\
  sbs\_03 &      0.076 & 0.066 &  7.104 & \tb{0.025} & \ul{0.028} \\
 spms\_01 &      0.210 & 0.173 & 20.253 & \tb{0.057} & \ul{0.060} \\
 spms\_02 &      0.336 &     x & 43.040 & \ul{0.200} & \tb{0.198} \\
 spms\_03 &      0.174 & 0.266 & 23.289 & \ul{0.062} & \tb{0.053} \\
  tnp\_01 &      0.090 & 0.221 &  1.467 & \tb{0.033} & \ul{0.041} \\
  tnp\_02 & \ul{0.110} & 0.779 &      x &          x & \tb{0.070} \\
  tnp\_03 &      0.089 & 0.382 &  1.051 & \ul{0.053} & \tb{0.042} \\
\hline\hline
\end{tabular}

\begin{tablenotes}
\small
\item *All values are in meter.
The best APEs among the methods are in \tb{bold}, the second best are \ul{underlined}. 'x' demotes a divergent experiment. FLIO is short for FAST-LIO2 SLICT2-2R refers to the number of point clouds that undergo re-associations in the inner loop.
\end{tablenotes}

\end{threeparttable}
\end{table}

\begin{table}
\centering
\renewcommand{\arraystretch}{1.25}
\begin{threeparttable}

\caption{APE of LO and LIO methods on MCD dataset.} \label{tab: mcd ate}


\begin{tabular}{|cccccc|}
\hline
\bf Sequence   &  \bf  FLIO & \bf CLIC & \bf CTICP &    \bf SLICT & \bf SLICT2-2R \\
\hline\hline
  xxx\_day\_01 &      0.901 & 25.194 &      x & \ul{0.570} & \tb{0.498} \\
  xxx\_day\_02 & \ul{0.185} &  0.503 &  0.807 &      0.188 & \tb{0.169} \\
  xxx\_day\_10 &      1.975 & 22.775 & 64.911 & \ul{0.842} & \tb{0.745} \\
xxx\_night\_04 &      0.902 &  4.154 & 69.024 & \ul{0.382} & \tb{0.344} \\
xxx\_night\_08 &      1.002 & 22.275 &      x & \tb{0.651} & \ul{0.817} \\
xxx\_night\_13 & \ul{1.288} & 21.299 &      x &      2.773 & \tb{0.631} \\
\hline
\end{tabular}

\begin{tablenotes}
\small
\item *All values are in meter.
The best APEs among the methods are in \tb{bold}, the second best are \ul{underlined}. 'x' demotes a divergent experiment.
\end{tablenotes}

\end{threeparttable}
\end{table}

In all experiments with SLICT2, we set knot length to 0.01s, the spline order is 4, the window size 3, the number of internal iterations 3, the re-associated clouds 2, and the maximum number of lidar factors is capped at 8000.
All experiments are carried out on a PC with intel core i9 CPU with 32 threads. Each method is run for 5 times and the best APE (Absolute Position Error) metric reported by the evo package \cite{grupp2017evo} for each sequence is recorded in \tref{tab: ntu ate} and \tref{tab: mcd ate}. 

\begin{figure}
    \centering
    \includegraphics[width=0.95\linewidth]{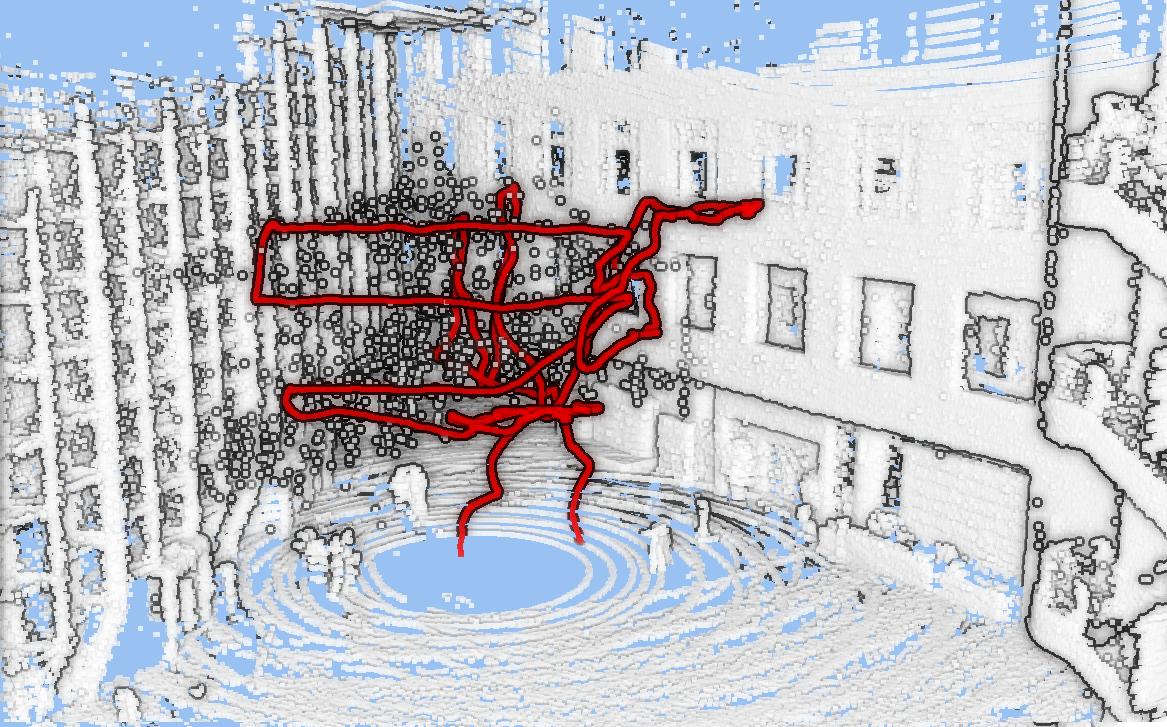}
\caption{The environment in the tnp\_02 sequence in NTU VIRAL dataset with significant noise from sun glare and glass wall. Also when the UAV is flying high (the trajectory of UAV is in red), there are few features that can constrain the vertical drift. This leads to the divergence of SLICT.}
\label{fig: ntu tnp 02}
\end{figure}

For the NTU VIRAL benchmark, we can see that SLICT2 has the best performance by the number of sequences with lowest APE, followed closely by SLICT. FAST-LIO2 and CLIC have somewhat similar performance, while CT-ICP has the lowest performance.
The close performance between SLICT and SLICT2 is expected. This is because SLICT2 shares similar frontend with SLICT, and the sequences are at relatively low speed, therefore the advantages of B-spline trajectory representation in SLICT2 is not yet evident over the PLI approach of SLICT. However we can see that SLICT2 does not diverge where SLICT does, which is due to the internal association measure which helps the method converge faster and more accurately in the challenging environment (Fig. \ref{fig: ntu tnp 02}). We also note that SLICT cannot run in real time even when only a 16-channel lidar is used.

{For the MCD benchmark, we have a more definitive result where SLICT2 has the smallest APE in the most sequences. At the second place is SLICT, followed by FLIO. Fig. \ref{fig: maps} shows the map and trajectory output of FAST-LIO and SLICT2 in one sequence. This demonstrates the benefit of the B-spline continuous-time formulation with dense number of control points, which can better capture the dynamics of high speed sequence and UAV motion.
On the contrary, CT-ICP, being a lidar-only method that uses PLI, struggles with the high speed sequences and can easily diverge in some long sequences. Finally, CLIC is slightly better than CT-ICP, however its performance is still not as good as FAST-LIO. It should also be noted that despite its relatively good performance, SLICT cannot run in real time even with the sparse livox point cloud. In addition, CT-ICP and CLIC cannot run in real time with such high number of lidar factors set for SLICT2, and CLIC frequently crashes when we try to set a dense number of control points like SLICT2.}

\subsection{Processing time} \label{sec: time analysis}

To study the efficiency of SLICT2, we carefully added multiple timers in the program to record the processing time for each step and sub-step in the xxx\_day\_01 sequence.

\begin{figure}
    \centering
    \begin{overpic}[width=\linewidth,
                   ]{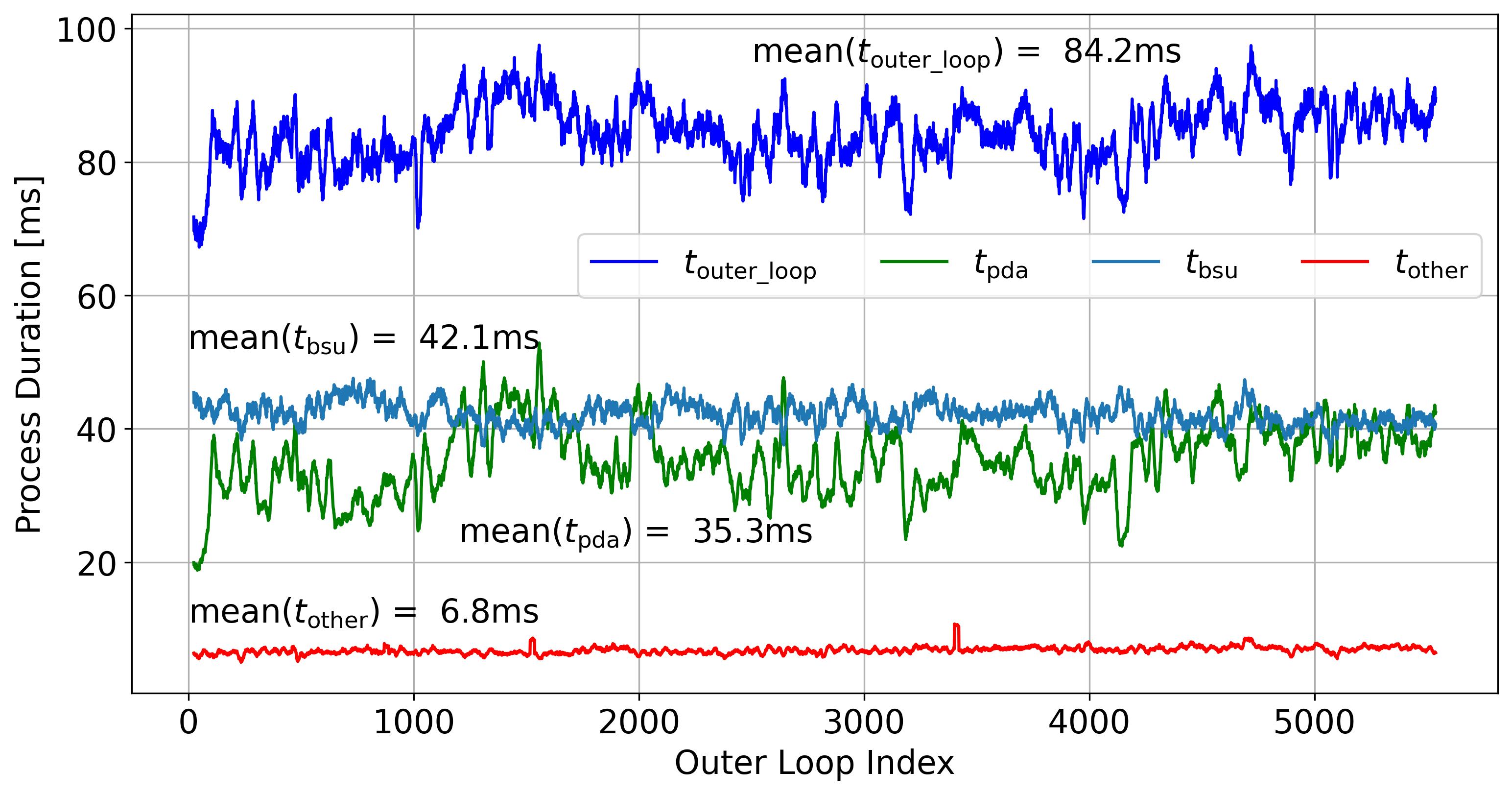}
    \end{overpic}
\caption{Processing time of the outer loop, the PDA process, the BSU process, and others. Note that the values have been smoothed out by applying a 20-step sliding window average.}
\label{fig: main time expenses}
\end{figure}

In reference to Fig. \ref{fig: main time expenses}, we define $t_\mathrm{outer\_loop}$ as the processing time of one outer loop (Fig \ref{fig: main pipeline}). Next, $t_\mathrm{pda}$ is defined as the total time for the three PDA iterations in the inner loops and one in the outer loop (step 2.2 and 2.1 in Fig. \ref{fig: main pipeline}). Similarly, $t_\mathrm{bsu}$ is the total time of three iterations of the BSU process in the inner loops. Finally, the $t_\mathrm{other}$ refers to other processes such as copying data, feature selection, visualization, making diagnostic reports etc.

From the plot, $t_\mathrm{loop}$ maxes at 100ms and the mean value is about 83ms, which confirms that SLICT2 can run in real time. Noteably, $t_\mathrm{pda}$ varies quite significantly, while $t_\mathrm{bsu}$ and $t_\mathrm{other}$ are relatively stable. This is due to the variation in size of the input point clouds, which has direct effect on the processing time for the down-sampling, deskew, and association processes. After the PDA process is completed, the optimization problem has a fixed size of 8000 lidar factors, thus $t_\mathrm{bsu}$ consumes more or less the same amount of time in every loop.

\begin{figure}
    \centering
    \begin{overpic}[width=\linewidth,
                   ]{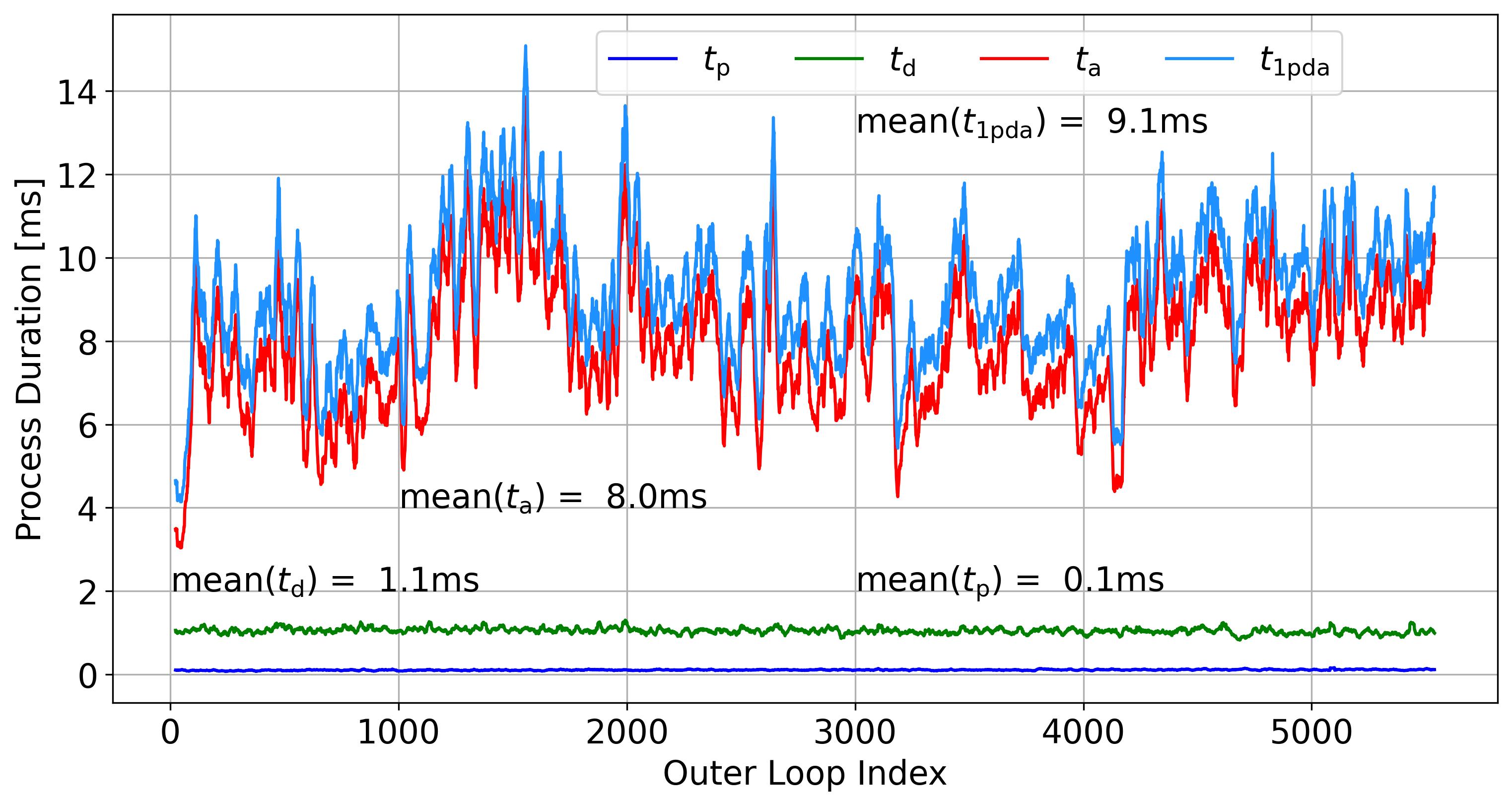}
    \end{overpic}
\caption{Processing time of one PDA iteration.}
\label{fig: pda time expenses}
\vspace{-0.4cm}
\end{figure}

Fig. \ref{fig: pda time expenses} shows the processing time of sub-processes of one PDA iteration, where $t_\text{1pda}$ refers to the PDA process time of the 1st inner iteration, and $t_\text{prop}$, $t_\text{desk}$, $t_\text{assoc}$ the processing time for propagation, deskew, and association respectively. We can see that the majority of processing time is expended on associating the lidar points with the map, while IMU propagation takes less than 1ms, and deskew less than 2ms.

\begin{figure}
    \centering
    \begin{overpic}[width=\linewidth,
                   ]{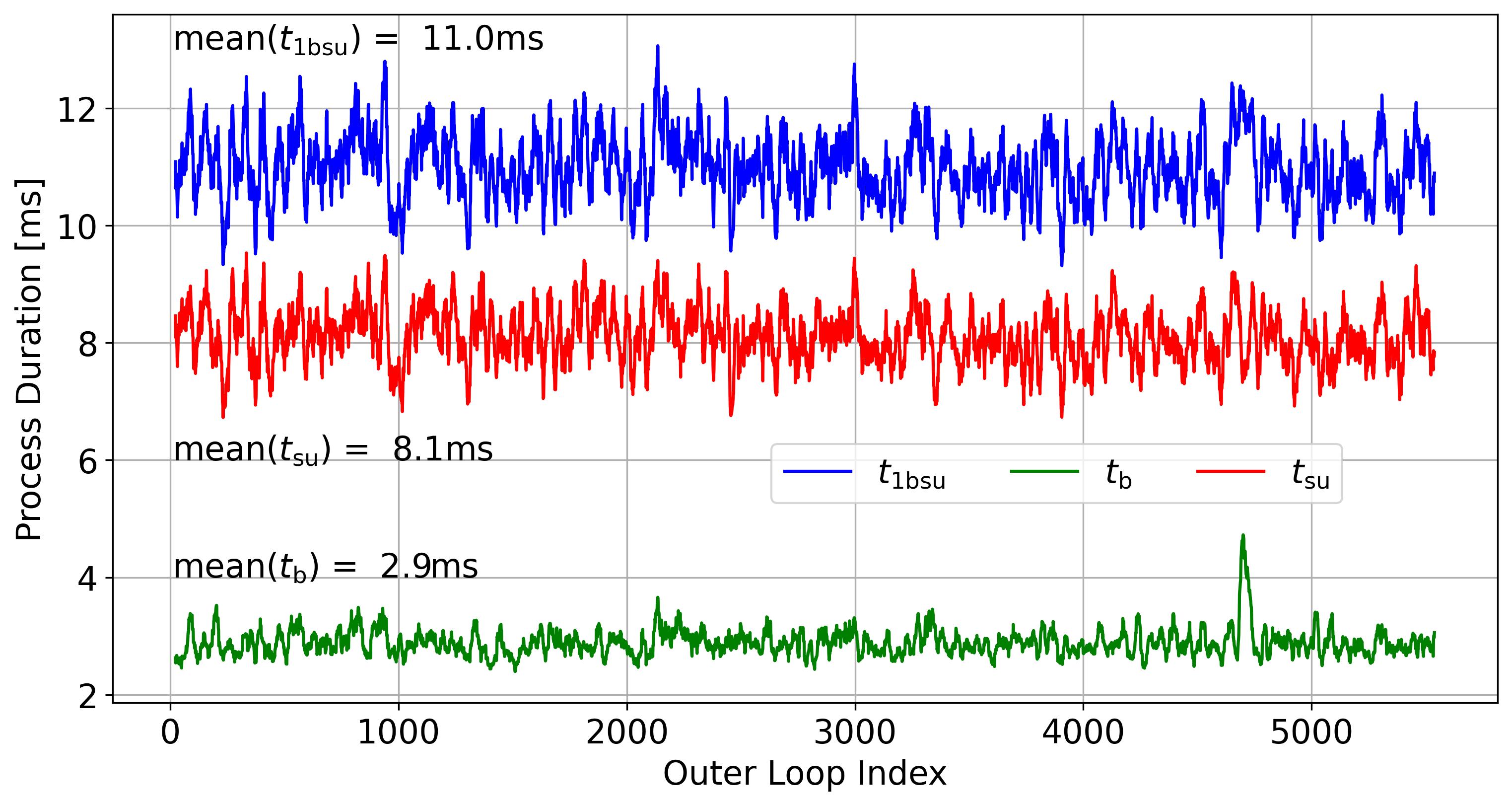}
    \end{overpic}
\caption{Processing time of one BSU iteration.}
\label{fig: bsu time expenses}
\vspace{-0.4cm}
\end{figure}

Similarly, we look into one BSU iteration in Fig. \ref{fig: bsu time expenses}. Here $t_\text{1bsu}$ is the processing time of the BSU sequence in the first inner loop. The processing of step 3.1 is denoted as $t_\text{b}$, and the processing time for step 3.2 and 3.3 are lumped together into $t_\text{su}$ as the update step is quite small compared to others.

It is clearly shown in Fig. \ref{fig: bsu time expenses} that the calculation of residuals and Jacobians of thousands of factors can be completed in a few milliseconds thanks to our parallelization scheme. On the other hand, the majority of time is spent on the solving step. This offers an insight into how the efficiency can be improved: since we only use a solver of the Eigen library for this task, a better solver or library can reduce the time spent on this process.

\begin{figure}
    \centering
    \begin{overpic}[width=\linewidth,
                   ]{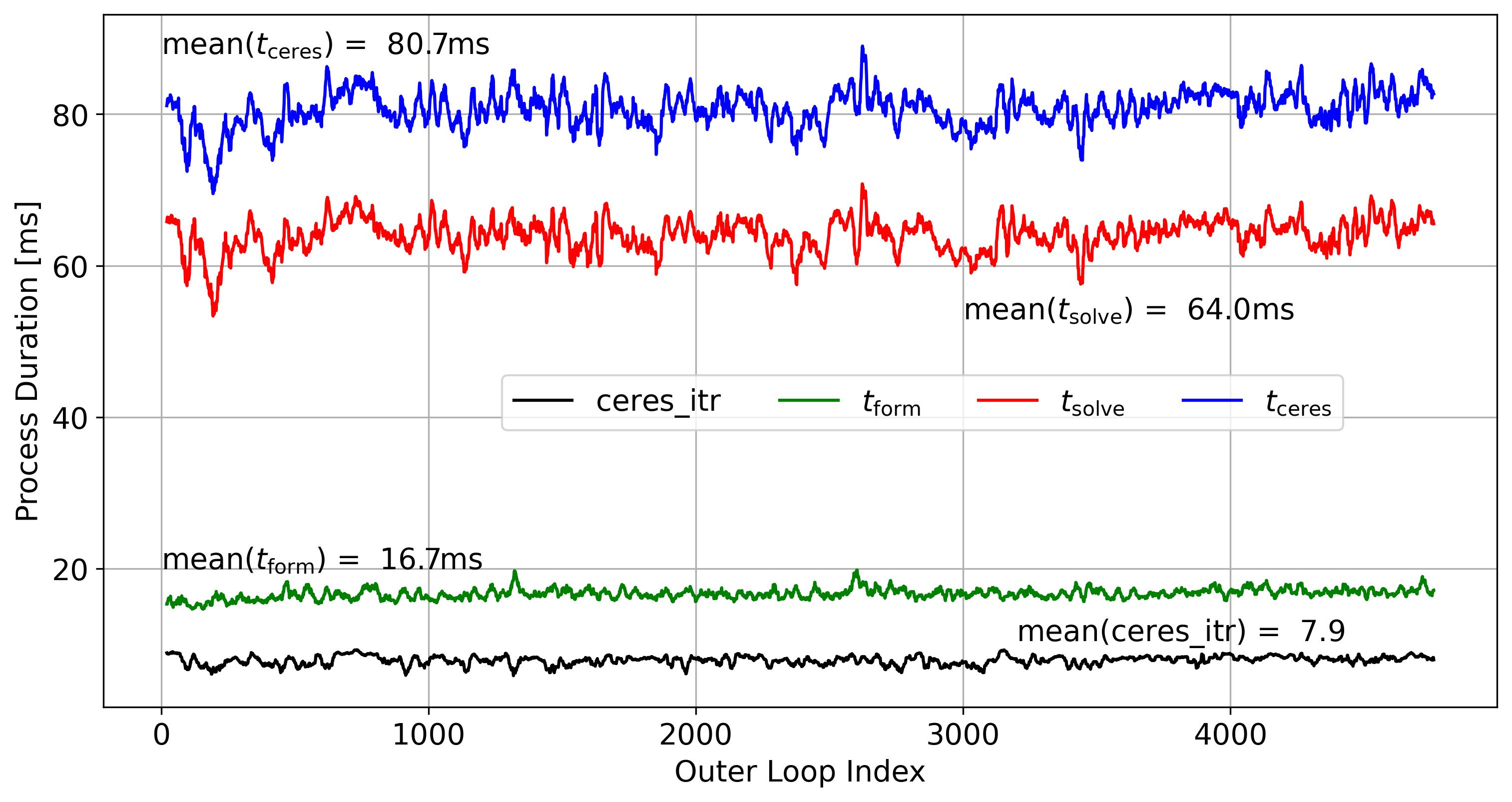}
    \end{overpic}
\caption{Processing time of one BSU iteration by using Ceres.}
\label{fig: Ceres bsu time expenses}
\end{figure}

Finally, to highlight the efficiency of the SLICT2 solver, we replace steps 3.1, 3.2 and 3.3 in our pipeline with a Ceres-based procedure, and record the time costs in Fig. \ref{fig: Ceres bsu time expenses}. In this plot, $t_\text{ceres}$ refers to the processing time to build the cost function and solve it with Ceres before moving to step 2.2, $t_\text{form}$ refers to the time needed to formulate a Ceres problem with 8000 lidar factors (plus a few hundred IMU factors), and $t_\text{solve}$ is the time needed to optimize the cost function. Here, $t_\text{solve}$ should be equivalent to $t_\text{1bsu}$ in Fig. \ref{fig: bsu time expenses}. We also report the number of internal iterations that Ceres takes before reaching the optimal solution, this is referred to as $\text{ceres\_iter}$.

First, we notice that $t_\text{form}$ alone is already longer than the whole BSU process when using SLICT2 solver. This is an overhead that is common for using NLS, not unique to Ceres. We also notice that the average time for Ceres to reach the optimal solution is 64ms, for about 7.9 iterations on average. If we divide 64ms by 7.9, we obtain the average time for an iteration of 8.1ms, which is close to the amount of time that SLICT2 needs for calculating the residual, Jacobian, and solving \eqref{eq: solution}. However, this is only one iteration, and Ceres needs up to 8 iterations to converge due to its internal checks to ensure monotonic decrease in each iteration. This again shows that the NLS-based approach is not as efficient as the method implemented in SLICT2.

\subsection{Ablation study}

As shown in Fig. \ref{fig: pda time expenses}, the association process takes the majority of time in the 2.2 step. To justify the iteration of this process in the inner loop, we need to show that it does improve the APE of the LIO scheme. To this end, we rerun SLICT2 with different number of point clouds undergoing re-association in the inner loop, denoted as $M$. Since the sliding window has a length of 3, $M$ is changed from 0 to 3 in these experiments.

From Tab. \ref{tab: ablation}, we can see that for each sequence, SLICT2-1 is generally improved over SLICT2-0, which is then improved further in SLICT2-2, and the effect does not distinctly improve with 3 re-associations. This shows that the internal association does offer advantageous effects on the accuracy of the CT-LIO system.

\begin{table}
\centering
\renewcommand{\arraystretch}{1.25}

\begin{threeparttable}

\caption{APE of SLICT2 with different re-associated point clouds.} \label{tab: ablation}


\begin{tabular}{|ccccc|}
\hline
 \bf Sequence  & \bf SLICT2-0 & \bf  SLICT2-1 & \bf  SLICT2-2 & \bf  SLICT2-3 \\
\hline\hline
  xxx\_day\_01 &    0.790 & \ul{0.527} & \tb{0.498} &      0.708 \\
  xxx\_day\_02 &    0.262 &      0.183 & \ul{0.169} & \tb{0.164} \\
  xxx\_day\_10 &    1.104 &      1.181 & \tb{0.745} & \ul{0.817} \\
xxx\_night\_04 &    0.670 &      0.466 & \tb{0.344} & \ul{0.354} \\
xxx\_night\_08 &    2.137 & \tb{0.706} & \ul{0.817} &      0.978 \\
xxx\_night\_13 &    2.960 &      0.771 & \ul{0.631} & \tb{0.477} \\
\hline
\end{tabular}

\begin{tablenotes}
\small
\item *All values are in meter. For each method, the best result out five runs is reported. The best APEs among the methods are in \tb{bold}, the second best are \ul{underlined}. 'x' demotes a divergent experiment. SLICT2-M indicates re-associations are applied to the last M point clouds on the sliding  window in the inner loop.
\end{tablenotes}

\end{threeparttable}

\end{table}

\section{Conclusion}  \label{sec: conclusion}

In this paper, we have developed a CT-LIO system using a basic yet efficient optimization approach, named SLICT2.
The key innovation in our method is the use of a simple solver which eliminates many computational overheads of conventional NLS, yet still achieves convergent result in only a few iterations. The performance is further enhanced by conducting association between each iteration. We also conduct thorough analysis on the processing time of each step in SLICT2's pipeline to gain insights into its efficiency.
The performance of SLICT2 is demonstratively competitive when compared to other SOTA methods, and 
it can be easily extended to address other issues such as online calibration, temporal offsets, and multi-modal sensor fusion. We believe SLICT2 has made a strong case for more efficient implementation of LIO systems in the future.



\balance
\bibliographystyle{IEEEtran}
\bibliography{references}

\end{document}